%% file: anonymous-submission-latex-2026.tex
\documentclass[letterpaper]{article} % DO NOT CHANGE THIS
\usepackage{aaai2026}  % DO NOT CHANGE THIS
\nocopyright % the AAAI copyright text is commented out in aaai2026.sty; this stops \maketitle from calling it (was an 'Undefined control sequence' error)
\usepackage{times}  % DO NOT CHANGE THIS
\usepackage{helvet}  % DO NOT CHANGE THIS
\usepackage{courier}  % DO NOT CHANGE THIS
\usepackage[hyphens]{url}  % DO NOT CHANGE THIS
\usepackage{graphicx} % DO NOT CHANGE THIS
\usepackage{natbib}  % DO NOT CHANGE THIS AND DO NOT ADD ANY OPTIONS TO IT
\usepackage{caption} % DO NOT CHANGE THIS AND DO NOT ADD ANY OPTIONS TO IT
\usepackage{algorithmic}

\usepackage{placeins} % KELSEY ADDED
\usepackage{afterpage} % KELSEY ADDED
\usepackage{longtable} % KELSEY ADDED
\newtheorem{defn}{Definition} % KELSEY ADDED
\usepackage{amssymb} % KELSEY ADDED

\usepackage{newfloat}
\usepackage{listings}
\usepackage{graphicx}
\usepackage{caption}
\usepackage[most]{tcolorbox}
\usepackage{multirow}
\usepackage{subcaption}

\usepackage{booktabs}

\usepackage{graphicx}
\usepackage{caption}
\usepackage[most]{tcolorbox}
\usepackage{multirow}
\usepackage{subcaption}
\usepackage{booktabs}
\usepackage[font=large]{caption}
\usepackage{pgfplots}
\pgfplotsset{compat=1.18}
\title{Do You Even Know What I Want? \\ Improving LLMs Ability to Detect Hidden User Intent}

\author {
    Kelsey Sikes\textsuperscript{\rm 1},
    Sarath Sreedharan\textsuperscript{\rm 1}
}
\affiliations {
    \textsuperscript{\rm 1}Department of Computer Science, Colorado State University, CO, USA\\
    ksikes@colostate.edu, sarath.sreedharan@colostate.edu
}

\usepackage{bibentry}
\input{commands}

\begin{document}

\maketitle

\input{AnonymousSubmission/LaTeX/paper-parts/0-abstract}
\input{AnonymousSubmission/LaTeX/paper-parts/1-introduction}

\input{AnonymousSubmission/LaTeX/paper-parts/3-background}
\input{AnonymousSubmission/LaTeX/paper-parts/4-running-example}

\input{AnonymousSubmission/LaTeX/5-evaluation}

\input{AnonymousSubmission/LaTeX/6-Discussion}

\input{AnonymousSubmission/LaTeX/paper-parts/2-related-work}

\input{AnonymousSubmission/LaTeX/8-conclusion-new}
\bibliography{aaai2026}
\onecolumn
\input{AnonymousSubmission/LaTeX/paper-parts/7-appendix}

\end{document}

%% file: commands.tex
\newcommand{\Action}[2]  {\ensuremath{\mathit{#1}^{#2}}}
\newcommand{\preP}[1]    {\ensuremath{\Action{pre}{#1}_+}}
\newcommand{\preN}[1]    {\ensuremath{\Action{pre}{#1}_-}}

\newcommand{\sharedPreP} {\preP{}}
\newcommand{\sharedPreN} {\preN{}}

%% file: AnonymousSubmission/LaTeX/paper-parts/0-abstract.tex
\begin{abstract}
As with conventional AI agents, a common source of misalignment between Large Language Models (LLMs) and their human users is misspecified goals. Often, this happens when the human has skewed beliefs about the system's model and provides it with information that causes behavior different from what they expected, with potentially unwanted side effects. To ensure goal alignment, one solution would be for the LLM to ask the human questions about their objectives. However, it has been well-documented that LLMs most often provide default responses to user requests rather than try and ask questions to clarify their true intent. In this paper, we address these challenges by adapting a novel formulation for the objective misspecification problem to work with LLMs. We then test if LLMs that are prompted to ask questions about a user's true underlying objectives are better at identifying them compared to those that do not. We evaluate our method using four state-of-the-art LLMs tested on five standard IPC domains, showing that for some domains, asking questions greatly improved model performance.
\end{abstract}

% However, it has been well-documented that LLMs most often default to answering user queries rather than autonomously asking questions about them. 

% However, it has been well-documented that LLMs most often provide default responses to user requests, rather than trying to ask question to clarify their true intent

%% file: AnonymousSubmission/LaTeX/paper-parts/1-introduction.tex
\section{Introduction}
A primary cause of value misalignment is misspecifed goals on the part of the human, often resulting from their incorrect beliefs about a robot's\footnote{We use the terms robot and agent interchangeably but these can refer to any autonomous agent, physically embodied or otherwise.} current state and capabilities. Reasons for this can be widespread, ranging from faulty reasoning to a human's own cognitive disabilities or lack of experience working with agents. When this happens, a human may partially specify their goals to an agent only to find that what it achieves varies drastically from what they expected it to achieve. In benign cases, this could be frustrating, but in others it might produce unwanted side effects that could threaten the human, a phenomenon commonly studied in AI Safety research \cite{christian_alignment_2021,russell2019human}.

Like traditional AI agents, Large Language Models (LLMs) can also fail to satisfy a user's goals when faced with misspecification. Here, a seemingly simple way to circumvent this would be for the LLM to ask the human clarifying questions about what they want, reducing any knowledge asymmetries. However, despite their many capabilities, LLMs often default to immediately answering user queries rather than eliciting additional information to provide more precise answers \cite{OpenAITechReport,Gemini2024}. Though some recent work has hypothesized why this is \cite{ModelingFutureConvos}, to be truly useful, such systems must develop the ability to identify a user's true objectives and reconcile any model differences inhibiting their achievement where possible.

\begin{figure}[t]
    \centering
    \includegraphics[width=0.48\textwidth]{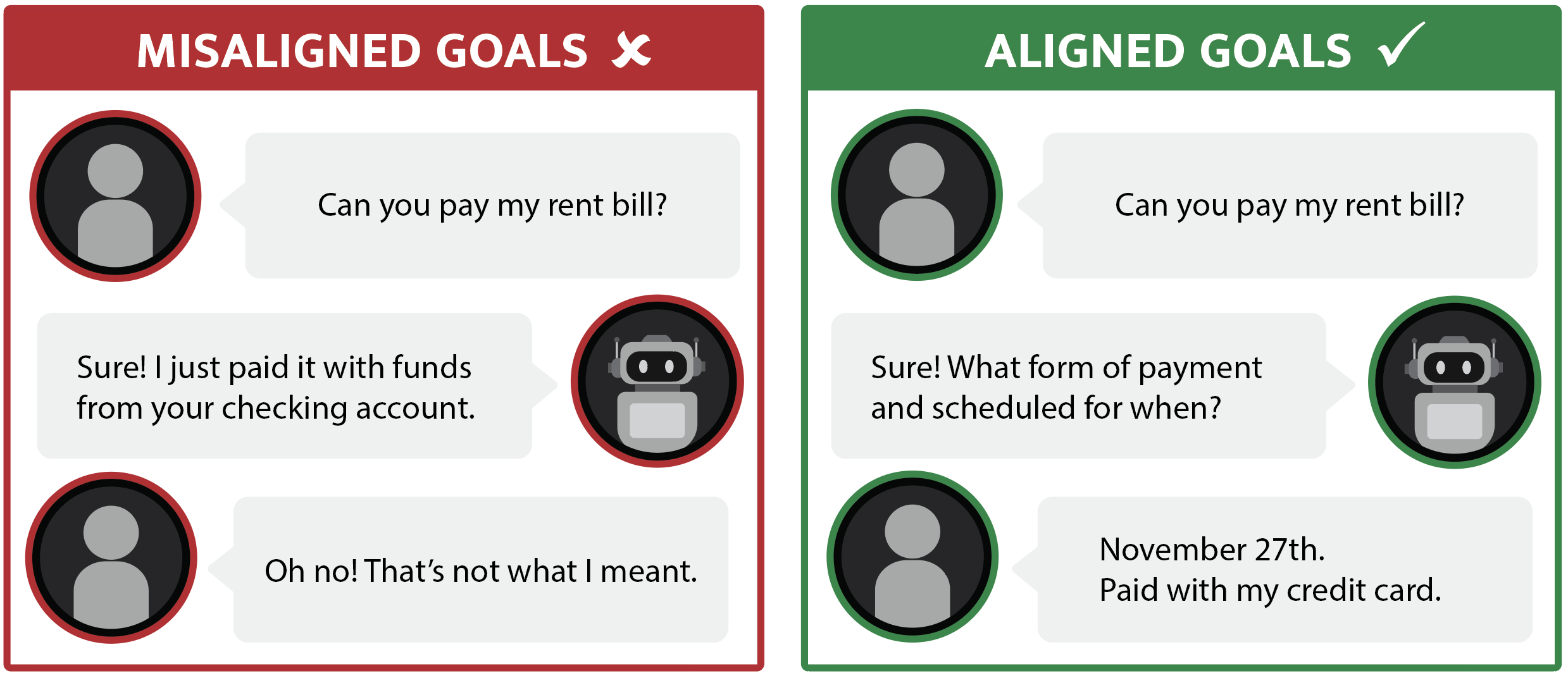}
    \caption{A depiction of the banking chatbot example. In the first scenario, a chatbot fails to properly satisfy a human's goals after they are misspecified based on incorrect beliefs about its model. In the second scenario, the chatbot asks the human clarifying questions about their goals to resolve any misalignments before successfully accomplishing them.} 
    \label{fig:goals_image}
\end{figure}

Recently, \citet{Mechergui_Sreedharan_2024} defined the problem of \textit{Human-aware goal alignment}, which formalizes the value alignment problem to account for differences between a user's expectations and an agent's decisions. They achieved this by extending the Human-Aware AI framework \cite{HumanAwareAIFramework}, which uses psychological concepts of mental models \cite{Premack_Woodruff_1978}, to handle goal misspecifation by querying users to learn more about their hidden objectives. In this paper, we adaptively apply this problem formulation to a separate human-robot goal misspecification problem that we are trying to solve: whether LLMs can reason about goal misalignment and identify good questions which resolve it. Thus, given an agent model and a human model, we test if LLMs can detect model discrepancies and proactively ask questions which reveal a user's true intent. We do this by empirically evaluating four state-of-the-art LLM’s on their ability to identify a user’s true goals, both out-of-the-box and after question’s have been asked them, using five standard IPC domains. We then quantify the effect of each model asking questions and analyze what factors led to correct versus incorrect goal classifications. We show that for some domains, asking questions substantially improved model performance.

%% file: AnonymousSubmission/LaTeX/paper-parts/3-background.tex
\section{Background}
To evaluate the ability of LLMs to determine a user's true goals we will rely on PDDL problems which are composed of two things: a domain and problem file. We will represent our problem as $\mathcal{M} = \langle \mathcal{D}, \mathcal{I}, \mathcal{G}\rangle$. Here, we will further define the domain $\mathcal{D}$ by the tuple $\mathcal{D} = \langle \mathcal{F, A}\rangle$, where $\mathcal{F}$ is a set of propositional fluents (i.e. boolean variables) that describe a potential state $s$, and $\mathcal{A}$ is a set of actions that upon execution change their state. We further define each action by the tuple $a = \langle \sharedPreP(a), \sharedPreN(a), add(a), del(a)\rangle$ which represents the action's positive/negative preconditions, and add/delete effects. An action can only be executed if its preconditions are satisfied, then its add and delete effects describe how it updates the state. We define the problem file $\mathcal{P}$ by the tuple $\mathcal{P} = \langle \mathcal{O, I, G}\rangle$ where $\mathcal{O}$ are objects, $\mathcal{I}$ is the initial state describing the true starting facts of the world, and $\mathcal{G}$ is a partial goal specification (i.e. $s\supseteq \mathcal{G}$). To move to a new state, we will use the transition function $\mathcal{T}$ where

\vspace{-2pt}
\[ \mathcal{T}(a, s, \mathcal{D}) = \begin{cases}
   (s \setminus del(a)) \cup add(a), \ ~\textrm{ if }  pre_{+}(a) \subseteq s \\
   \textrm{\em{undefined} } ~~~~~~~~~~~~~~~~~~~~~~~\textrm{otherwise}
\end{cases}\]

At each state, $\mathcal{T}$ captures the effects of executing an action. To work with with action sequences, we will overload this notation such that $\mathcal{T}(\langle a_1, \ldots, a_k \rangle, s, \mathcal{D}) = \mathcal{T}(a_1,\, \mathcal{T}(\langle a_2, a_3, \ldots, a_k \rangle, s, \mathcal{D}), \,\mathcal{D}).$ A plan $\pi$ is a sequence of actions which is also a solution to a planning problem if its execution from an initial state results in a goal state. Each action in a plan has a cost $c$, which we will assume is a unit cost. In our setting, a plan is optimal if it achieves a goal in the fewest actions possible. In a lifted domain all predicates contain variables. In our work we will use grounded domains in which all variables are replaced with compatible objects. For example, in the Blocksworld domain, the predicate \texttt{(on\ ?x\ ?y)} becomes \texttt{(on\_a\_b)} where the variables \texttt{?x} and \texttt{?y} are replaced with the block objects \texttt{a} and \texttt{b}.

%% file: AnonymousSubmission/LaTeX/paper-parts/4-running-example.tex
\section{Running Example}
Consider a situation where a bank has employed an online chatbot to help users manage their finances (shown in Figure ~\ref{fig:goals_image}). Now imagine an interaction where a human asks the chatbot to pay their rent, expecting payment to be scheduled for end of month using a credit card. Here, the user may have assumed the chatbot knew what they wanted or forgotten to fully specify their goals. Regardless, the chatbot chooses the optimal plan and immediately uses all money in their checking account to pay the bill. This triggers overdraft fees which the human now has no way to pay. Now, consider an alternate scenario where the chatbot assumes the human's goal specification may be incomplete, and therefore queries them to try and verify correctness. This prompts the human to provide additional details about their objectives, like an explanation of the plan they believed the chatbot would execute to achieve them. This provides the chatbot with more insight into the human's true underlying goals, while also revealing facts about the world it may have been unaware of. For example, that the human's rent bill is always paid with a credit card or that its due date changed. With this added understanding, the chatbot updates it's beliefs about the human's true goals, then satisfies them while avoiding unwanted negative side effects. The question now becomes how to generate such agent behavior.

%% file: AnonymousSubmission/LaTeX/5-evaluation.tex
\section{Goal Alignment Problem with LLMs}
The central focus of this work is to determine whether LLMs can ask questions that allow them to identify a user's true underlying goals. To simulate a scenario where this is necessary we will later provide each LLM with information to ask about: mismatched human-robot models and a human plan. Therefore, we will start by defining what these things are. First, we will denote the human and robot domain models, consisting of fluents and actions, by $\mathcal{D^H} = \langle \mathcal{F}, \mathcal{A^H}\rangle$ and $\mathcal{D^R} = \langle \mathcal{F}, \mathcal{A^R}\rangle$. The robot's model of the task is then $\mathcal{M^R} = \langle \mathcal{D^R}, \mathcal{I^R}, \mathcal{G^R}\rangle$ where $\mathcal{I^R}$ is the robot's initial state and $\mathcal{G^R}$ its goal specification. Recall that, in our problem, the human's beliefs about the robot's current state and capabilities do not match true reality. To reflect this incorrect understanding, we represent the human's belief model of the robot as $\mathcal{M^H} = \langle \mathcal{D^H}, \mathcal{I^H}, \mathcal{G^H}\rangle$ where $\mathcal{I^H}$ is the initial state the human believes the robot is starting from, and $\mathcal{G^H}$ is the human's specified goals. Note that in our problem setting the human and robot have different initial states but we make the common assumptions that they share the same fluents and that we have access to their models. In scenario 1 of the banking example, $\mathcal{G^H}$ would correspond to the human's rent bill being paid. However, remember that this resulted in the unexpected outcome of the robot immediately paying the bill with funds from their checking account. Thus, despite their specification, the human's true underlying goal was actually for the bill's payment to be scheduled for a later date using a credit card. This means $\mathcal{G^H}$ only reflects the human's specified goals based on their beliefs about the robot, not the human's true underlying goals. To represent the goals directly specified to the robot and which the human implicitly assumed it knew to achieve, we will use $\mathcal{G^*}$. To show this more precisely, we adopt the definition of a goal-misalignment problem from \citet{Mechergui_Sreedharan_2024}:

\begin{defn}
\textit{A goal specification is said to be misaligned with the human goal $\mathcal{G^*}$ for a robot domain model $\mathcal{D^R}$ and initial state $\mathcal{I^R}$, if there exists an action sequence $\pi = \langle a_1,...,a_k \rangle$ such that $\mathcal{T(\pi, I^R, D^R) \supseteq G^H}$, but $\mathcal{T(\pi, I^R, D^R) \nsupseteq G^*}$}
\end{defn}

To avoid this misalignment, we will provide the LLM with more information about the human's goals by giving it their plan $\pi^\mathcal{H}$. This corresponds to the action's the human believes the robot will execute to achieve their specified goals therefore providing valuable insight into what they could be. However, such actions may not be executable in the robot's model or achieve the final goal state the human imagined. For example, if the human wanted their rent bill to be paid with a credit card but the robot only had access to their checking account, a goal state including this fluent would be unachievable. While important, we consider this orthogonal to our problem, which is only focused on the LLM identifying all true human goals $\mathcal{G^*}$. However, such differences are something the LLM must reason about when determining which facts are relevant to the human's true goals. Hence given the human's plan, we then know that $\mathcal{G^*}$ is some subset of the final goal state they expected. The LLMs task is then to figure out what this group of goals is, given the fact that for our instantiation of this problem the $\mathcal{G^R} \subseteq \mathcal{G^*}.$ For this we will have the LLM ask an unbounded number of questions to another LLM approximating a human which has more information about the human's goals and beliefs about the task. Using this information, the mismatched human-robot models, and the human's plan, the LLM will then determine what the human's true underlying goals are. A solution to this goal alignment problem is then when the LLM correctly identifies these predicates and the robot model can be updated such that $\mathcal{G^R} =\mathcal{G^*}$. 

\section{Evaluation}
We evaluated the ability of four state-of-the-art LLMs to ask questions about a user's underlying goals and then identify them given a true task model and a human belief model of the task. We did this by conducting two experiments. In the first, single-step prompting was used to directly ask each model what the human's goals were. In the second, multi-stage prompting was used to have each model ask questions about the human's goals, and then, using the returned response from a fifth LLM simulating a human, identify them. We include all prompts and experimental details, including our full results, in the appendix of this work.

\subsection{Dataset} We used a public problem instance generator \cite{seipp-et-al-zenodo2022} to create 500 International Planning Competition (IPC) problem variations \cite{IPC2000,IPC2002}. This spanned the Blocksword, Childsnack, Driverlog, Rovers, and Satellite domains for which 100 instances of each were made. From these, all action costs were removed and each goal was modified to include a varied set of 2-6 predicates. All instances were then duplicated to create human-robot model pairs and grounded. In all robot models, 1-5 goal predicates were randomly removed (this never exceeded the original number of goal predicates). From the human models, 2-4 random predicates were added to the initial state (excluding those already included in the goal). All such modifications were randomly determined and only accepted if they resulted in a unique problem instance for which a valid human/robot plan could be generated. This created misaligned models with different initial states and goals, representative of a human who expects a robot to achieve certain goals based on their incorrect beliefs about its knowledge and capabilities. Using the Fast Downward Planner with A* search and the Landmark Cut heuristic \cite{FastDownward} optimal plans for both grounded models were generated. For a human-robot model pair to be kept, the human's plan had to be valid only in the human's model, and shorter than the one produced by their original model (i.e. the plan generated prior to initial state predicates being added). Otherwise, the model pair was discarded and new problem variations were created until one which met this criteria was found. VAL \cite{val_github} was used to check which models a plan was valid in. 

\subsection{Prompts \& Inputs} We created six prompts to test LLMs ability to identify, ask, or answer questions about a human's true goals. Inputted into all prompts (except Prompt 4) were the human's plan, human/robot domain file, robot problem file, and human problem file, but with the robot's goals. Here, only the robot's goals were given, so the LLM had to reason about what hidden goals the human may have that the robot was unaware of. To minimize context window issues, and because the human and robot used the exact same domain file, this was inputted once. Inputted into Prompt 4 were the human's true goals and questions about them. Prompts 1 and 2 were used in Experiment \#1. The purpose of these was to test LLMs ability to identify hidden user intent by directly asking each model to name a human's true goals. All other prompts were used in Experiment \#2. Here, Prompt 3 was created to prompt each LLM to ask questions about a user's intent. Prompt 4 was made to answer questions about a user's intent. This prompted an LLM to answer questions about a human's goals generated using Prompt 3. Because experimenting with a human would be too expensive, this LLM was meant to simulate a person with full knowledge of the goal providing information about it. Prompts 5.1 and 5.2 also tested a model's ability to identify hidden user intent by asking for all human goals. However, these included answers to the questions from Prompt 4 to help with this. 
 
\subsection{Setup} Due to their generally strong reasoning abilities and large context window sizes, Gemma-4-26B, Llama-3.3-70B-Instruct, Mistral-Small-24B-Instruct, Qwen3.5-27B, and Deepseek-R1-Distill-Llama-70B\footnote{For succinctness we refer to these as Gemma, Llama, Mistral, Qwen, and Deepseek from this point on.} were used in our experiments. To ensure all outputs were deterministic so a baseline level of performance could be established, all models were set to a temperature of 0. Since we are unaware of any existing work that addresses this specific problem, our metrics focused on what percent of human goal predicates (only hidden and all potential predicates) each model was able to identify, and the number of correct problem instances this translated into. All experiments were run with a time limit of 20 minutes/instance and executed using an H100 GPU. Each model was downloaded from hugging face and run locally or accessed using an API. 

\input{AnonymousSubmission/LaTeX/paper-parts/SMALL_identify-preds-table}

\subsection{Experiment \#1: Identifying Hidden Predicates}
This experiment evaluated LLMs ability to ask about a human's true goals without explicit instruction and then determine them given mismatched human-robot models and the human's plan. The motivation for this was twofold; first we wanted to see if the LLMs would ask clarifying questions unprompted about the human's objectives. Then, irrespective of any questions asked, we wanted to quantify how accurate each LLM was at identifying the human's goals. To accomplish this, Prompt 1 and 2 were used which directly asked each LLM to identify the human's true goals using the same inputs but with slightly different language. During separate runs, these were iteratively updated with problem-specific input (i.e. human plan, human/robot problem files, etc.) for all 100 domain instances. These updated prompts were then fed to the Gemma, Llama, Mistral, and Qwen models which were asked to identify all human goal predicates in a single-shot. Because the robot's goals are a subset of the human's true goals, this meant naming all goal predicates hidden from and originally given to the robot. To illustrate this, for the Blocksworld domain, imagine the human's true goals were $(on\_b5\_b1)$, $(holding\_b2)$, and $(on\_b3\_b4)$, but the robot was only given the goal of $(on\_b5\_b1)$. For the instance to be correct, as its final answer the model would have to identify the goal already given to the robot $(on\_b5\_b1)$, as well as the hidden goals it was unaware of, $(on\_b3\_b4)$ and $(holding\_b2)$.

\input{AnonymousSubmission/LaTeX/paper-parts/added-preds-table}

\noindent\textbf{Results} 
Table~\ref{identify-preds-SMALL} shows results for the Gemma, Qwen and Mistral models, as well as the total number of problem instances processed for each. Due to timeouts/context window issues, a large portion of the Mistral and Llama instances did not complete. We include our full results in the appendix, primarily comparing results from the other models here for this reason. As expected, none of the LLMs asked questions about the provided inputs, so we also omit this information. As mentioned, the provided prompts asked each model to identify all true human goals as their answer, however, this was met with varied success. For this reason, to determine the type and number of predicates each model was able to identify results were organized into All and MissH categories. The All column represents the total percentage of all possible goal predicates (i.e. hidden and originally given) that a model was able to identify. Similarly, the MissH column represents the total percentage of hidden goal predicates that were withheld from the agent but part of the human's true goals that were identified. The All \% and MissH \% Correct columns show the percentage of correct problem instances for each scoring criteria. To be correct, an answer had to include every predicate from either the full or hidden set of human goals, with no extraneous predicates added.

Prompt 1 identified the greatest number of goal predicates for the All and MissH categories while also achieving the highest percentage of \% Correct instances for both. In general, Qwen exhibited the strongest model performance, most notably identifying 88\% All and 75\% MissH goal predicates in the Rovers domain. Across domains, this translated to the highest percentage of All and MissH \% Correct instances at 52\% and 56\% respectively. For the Satellite domain, Gemma exhibited the worst performance, only identifying 47\% All and 30\% MissH goal predicates. This translated to the lowest number of MissH \% Correct instances at 8\%. While Mistral had the lowest number of All \% Correct instances for the Blocksworld domain with 3\%. 

Nevertheless, for both prompts, the average All and MissH \% Correct instances was well under 50\%. This indicates that a model could identify a large number of goal predicates in either category, yet still get many answers incorrect. Such results also show that a grounded domain's number of actions and predicates had little impact on correctness. To illustrate this, each Blocksworld instance contained an average of 60 predicates and 93 actions. Yet, for Prompt 1, Gemma only got 13\% and Qwen 19\% of All \% instances correct. In contrast, each Rovers instance included an average of 194 predicates and 1071 actions. Yet for Prompt 1, Gemma got 22\% and Qwen 52\% of All \% instances correct. One possible explanation is the average number of extraneous predicates added to each model's answers, shown in Table~\ref{num-preds-added-to-answer-exp1}. For example, while Qwen added an average of 1 extra predicate to 25\% of its answers, for the Rovers domain, it only did this to 5\% of its answers. This implies that while incorrect answers for some domains may have stemmed from too few goal predicates, for others they may have resulted from too many.

\input{AnonymousSubmission/LaTeX/paper-parts/SMALL_identify-preds-experiment2}

\subsection{Experiment \#2: Asking Questions to Reveal Intent}
The purpose of this experiment was to evaluate whether the average model identification accuracy improved after questions were asked about a human's true goals. Here, we wanted to explore the quantity and type of questions asked, as well as the usefulness of the information elicited. For comparison purposes, the human-robot model pairs with discrepancies and human plan from Experiment \#1 were again used. To ensure most instances completed, only the Gemma, Qwen, and Mistral models were tested. This experiment was executed in three parts; First, each model was prompted to ask questions about the human's true goals using Prompt 3. Like before, this prompt was iteratively updated with problem-specific input (i.e. human plan, human/robot problem files, etc.) for all 100 domain instances. Next, these questions and the human's true goals were inputted into Prompt 4 and given to Deepseek. The model was then asked to answer these questions given the human's goals, approximating how another human might answer. Finally, Deepseek's answers were inputted into Prompts 5.1 and 5.2, again with problem-specific input for each instance. These were then fed back to each model separately, and used to directly ask each LLM to identify the human's true goals using slightly different language. Here, it is important to note that Prompt 5.1 used the exact same language from Prompt 1 but included Deepseek's answers, while Prompt 5.2 did the same for Prompt 2.

\noindent\textbf{LLM Questions}
Mistral averaged the most questions asked/model in response to Prompt 3 with 6, followed by Qwen with 4, and Gemma with 2. Because all questions related to the human's goals in some way, we organized these into two main categories based on content:

\begin{itemize}
    \item \textbf{Goal Clarification:} These tried to clarify what the human's true goals were independent of any plan by explicitly asking what they were, what objects/predicates were a part of them, if the LLMs intended answer was correct, etc. For example, a question from the Driverlog domain was, ``Does the human want the robot to move all packages or just some of them?"
   \item \textbf{Plan vs. Goal Alignment:} These addressed discrepancies between the robot's goals and the human's plan asking if the human plan revealed goals not present in the robot model, why it contained actions nonessential to the stated goals/inexecutable given the robot's initial state, etc. One such question from the Rovers domain was, ``Is capturing and communicating an image of objective 1 intended, given that the plan includes calibrating for it?"
\end{itemize}

These were auto-categorized using common word combinations/phrases indicating their type, aside from a few which required manual inspection. Across models most questions were focused on better understanding the robot's goals given the human's plan and its relationship to the provided human-robot models. This was most obvious for Qwen which focused less than 20\% of its questions on purely clarifying goals (shown in Appendix, Figure 8).

\begin{table*}[t]
\centering
\renewcommand{\arraystretch}{1.15}
\setlength{\tabcolsep}{6pt}

\begin{tabular}{llcccccc}
\toprule
\multirow{2}{*}{\textbf{Model}} & \multirow{2}{*}{\textbf{Domain}}
& \multicolumn{3}{c}{\textbf{Prompt 5.1}}
& \multicolumn{3}{c}{\textbf{Prompt 5.2}} \\ 
\cmidrule(lr){3-5} \cmidrule(lr){6-8}
& 
& \textbf{\# Inst.} & \textbf{\% Added Preds} & \textbf{Avg. \#}
& \textbf{\# Inst.} & \textbf{\% Added Preds} & \textbf{Avg. \#} \\ 
\midrule

\multirow{5}{*}{Gemma-4-26B}
& Blocksworld & 93 & 38\% & 2 & 90 & 26\% & 2 \\
& Childsnack  & 79 & 39\% & 2 & 82 & 35\% & 2 \\
& Driverlog   & 74 & 23\% & 2 & 79 & 25\% & 1 \\
& Rovers      & 61 & 44\% & 2 & 59 & 36\% & 2 \\
& Satellite   & 78 & 21\% & 2 & 77 & 16\% & 1 \\
\midrule

\multirow{5}{*}{Qwen3.5-27B}
& Blocksworld & 100 & 38\% & 2 & 100 & 45\% & 2 \\
& Childsnack  & 88 & 51\% & 2 & 88 & 53\% & 3 \\
& Driverlog   & 87 & 41\% & 2 & 87 & 41\% & 2 \\
& Rovers      & 65 & 34\% & 2 & 65 & 40\% & 2 \\
& Satellite   & 98 & 19\% & 1 & 97 & 27\% & 1 \\
\midrule

\multirow{1}{*}{Mistral-Small-24B}
& Blocksworld & 95 & 88\% & 3 & 96 & 78\% & 2 \\
\bottomrule

\end{tabular}
\caption{Percent of Experiment \#2 answers with extraneous goal predicates added and the average number.}
\label{num-preds-added-to-answer}
\end{table*}

\noindent\textbf{Deepseek Answers} In response to Prompt 4, Deepseek generally answered all questions, but its output quality varied. Because many responses included compound answers, categorizing these was difficult. We therefore provide some general observations instead. First, Deepseek's answers commonly included clarifying the human's objectives by explicitly listing every goal predicate. Alternatively, sometimes it would only list some goals predicates or the objects contained within them. Next, it would often explain why certain actions had been executed in the human's plan and what actions the robot should take as a result. Other times Deepseek would indicate the human's plan was a suggestion rather than a strict sequence that must be followed, sometimes going so far as to specify which predicates/actions the robot should focus on should this plan be inexecutable in its model. Most notable was the incorrect or misguided responses Deepseek would sometimes give. Though less frequent, the model would at times respond to an LLMs questions with exactly what it had asked, Yes/No Answers, or suggestions about what questions it thought the LLM should have asked. More widespread was its tendency to introduce incorrect information into otherwise correct responses. For example, making statements that were not factually correct like, ``Holding b2 allows the robot to manipulate other blocks without obstruction." Or by making claims like ``The goal list provided in the problem file is exhaustive." Occasionally, it gave answers based only on the human's goals, forgetting that the robot did not have access to them, with phrases like ``No, there are no additional goal states beyond what was explicitly listed in the human’s goals." Lastly, some response labels did not match the provided information. For example, using the label ``Clarifying Implicit Constraints" but then proceeding to discuss goal predicates.

\noindent\textbf{Results} Prompts 5.1 and 5.2 asked each model to identify a human's true goals using Deepseek answers about them. Table~\ref{identify-preds-SMALL-exp2} shows results for each model using the same scoring criteria from Experiment \#1. Surprisingly, Prompt 5.2 outperformed Prompt 5.1 in 70\% of the domains tested in the All \% Correct category. Of particular note here was the 47\% of Blocksworld instances it got correct using Gemma-4-26B. This far exceeded the previously best-achieved result using Prompt 1 by approximately 28\%. However, for the All, MissH, and MissH \% categories, Prompt 5.1 produced better results though by a relatively small margin. Such findings are significant because whereas a large performance gap existed in Experiment \#1 between Prompts 1 and 2, results for Prompts 5.1 and 5.2 showed much less variation with each outperforming the other in different categories. Recall that here the only difference between these sets of prompts was that in Experiment \#2 answers to each model's questions from Deepseek were included. This illustrates the fact that in some cases this additional information greatly improved model performance, especially when comparing Prompt 2 results to Prompt 5.2. Despite this, Prompt 1 still performed best overall. A possible reason for this is that Experiment \#2 saw a reduction in the average number of extraneous predicates added, but a significant increase in the percentage of answers to which they were added (Table~\ref{num-preds-added-to-answer}). For Prompt 5.1, Gemma added these to 33\% of answers, and Qwen to 36\%. For Prompt 5.2, Gemma added extra predicates to 27\% of answers, and Qwen to 41\%. In contrast, for Prompt 1, on average Gemma added extra predicates to 17\% of answers, and Qwen to 11\%. Lastly, in both experiments the Satellite domain consistently had the lowest percentage of All and MissH \% Correct instances by a large margin. This domain also had the greatest discrepancy between the number of true human versus originally given robot goals. This indicates that while the number of domain actions/predicates may not have effected model performance, the difference between these sets might have.

% \begin{table*}[t]
% \centering
% \renewcommand{\arraystretch}{1.15}
% \setlength{\tabcolsep}{6pt}

% \begin{tabular}{llcccccc}
% \toprule
% \multirow{2}{*}{\textbf{Model}} & \multirow{2}{*}{\textbf{Domain}}
% & \multicolumn{3}{c}{\textbf{Prompt 5.1}}
% & \multicolumn{3}{c}{\textbf{Prompt 5.2}} \\ 
% \cmidrule(lr){3-5} \cmidrule(lr){6-8}
% & 
% & \textbf{\# Inst.} & \textbf{\% Added Preds} & \textbf{Avg. \#}
% & \textbf{\# Inst.} & \textbf{\% Added Preds} & \textbf{Avg. \#} \\ 
% \midrule

% \multirow{5}{*}{Gemma-4-26B}
% & Blocksworld & 93 & 38\% & 2 & 90 & 26\% & 2 \\
% & Childsnack  & 79 & 39\% & 2 & 82 & 35\% & 2 \\
% & Driverlog   & 74 & 23\% & 2 & 79 & 25\% & 1 \\
% & Rovers      & 61 & 44\% & 2 & 59 & 36\% & 2 \\
% & Satellite   & 78 & 21\% & 2 & 77 & 16\% & 1 \\
% \midrule

% \multirow{5}{*}{Qwen3.5-27B}
% & Blocksworld & 100 & 38\% & 2 & 100 & 45\% & 2 \\
% & Childsnack  & 88 & 51\% & 2 & 88 & 53\% & 3 \\
% & Driverlog   & 87 & 41\% & 2 & 87 & 41\% & 2 \\
% & Rovers      & 65 & 34\% & 2 & 65 & 40\% & 2 \\
% & Satellite   & 98 & 19\% & 1 & 97 & 27\% & 1 \\
% \midrule

% \multirow{1}{*}{Mistral-Small-24B}
% & Blocksworld & 95 & 88\% & 3 & 96 & 78\% & 2 \\
% \bottomrule

% \end{tabular}
% \caption{Percent of Experiment \#2 answers with extraneous goal predicates added and the average number.}
% \label{num-preds-added-to-answer}
% \end{table*}

%% file: AnonymousSubmission/LaTeX/paper-parts/SMALL_identify-preds-table.tex
% \clearpage
% \afterpage{
% \onecolumn
% \begin{table*}[t]

% \clearpage
% \begin{table}[t]
% \begin{table}[p]
\begin{table*}[t]
\centering
\renewcommand{\arraystretch}{1.15}
\setlength{\tabcolsep}{6pt}

% -------- Prompt 1 --------
\begin{tabular}{llccccc}
\toprule
\multirow{2}{*}{\textbf{Model}} & \multirow{2}{*}{\textbf{Domain}}
& \multicolumn{5}{c}{\textbf{Prompt 1}} \\ 
\cmidrule(lr){3-7}
& 
& \textbf{\# Inst.} & \textbf{All} & \textbf{All \% Correct} & \textbf{MissH} & \textbf{MissH \% Correct} \\ 
\midrule

\multirow{5}{*}{Gemma-4-26B}
& Blocksworld & 95 & 65\% & 13\% & 33\% & 18\% \\
& Childsnack  & 85 & 71\% & 20\% & 49\% & 24\% \\
& Driverlog   & 82 & 63\% & 31\% & 37\% & 32\% \\
& Rovers      & 63 & 62\% & 22\% & 48\% & 29\% \\
& Satellite   & 92 & 47\% & 4\% & 30\% & 8\% \\
\midrule

\multirow{5}{*}{Qwen3.5-27B}
& Blocksworld & 100 & 72\% & 19\% & 44\% & 27\% \\
& Childsnack  & 87 & 74\% & 24\% & 52\% & 25\% \\
& Driverlog   & 89 & 82\% & 39\% & 66\% & 45\% \\
& Rovers      & 66 & 88\% & 52\% & 75\% & 56\% \\
& Satellite   & 98 & 74\% & 12\% & 55\% & 15\% \\
\midrule

\multirow{1}{*}{Mistral-Small-24B}
& Blocksworld & 100 & 63\% & 3\% & 33\% & 18\% \\
% & Satellite   & 39 & 79\% & 3\% & 77\% & 49\% \\
% \midrule
% \bottomrule
\end{tabular}

% \vspace{1em}

% -------- Prompt 2 --------
\begin{tabular}{llccccc}
\toprule
\multirow{2}{*}{\textbf{Model}} & \multirow{2}{*}{\textbf{Domain}}
& \multicolumn{5}{c}{\textbf{Prompt 2}} \\ 
\cmidrule(lr){3-7}
& 
& \textbf{\# Inst.} & \textbf{All} & \textbf{All \% Correct} & \textbf{MissH} & \textbf{MissH \% Correct} \\ 
\midrule

\multirow{5}{*}{Gemma-4-26B}
& Blocksworld & 97 & 64\% & 11\% & 28\% & 16\% \\
& Childsnack  & 85 & 69\% & 17\% & 46\% & 20\% \\
& Driverlog   & 87 & 61\% & 16\% & 28\% & 16\% \\
& Rovers      & 62 & 70\% & 13\% & 46\% & 29\% \\
& Satellite   & 93 & 52\% & 1\% & 25\% & 3\% \\
\midrule

\multirow{5}{*}{Qwen3.5-27B}
& Blocksworld & 100 & 69\% & 18\% & 37\% & 22\% \\
& Childsnack  & 87 & 71\% & 20\% & 46\% & 21\% \\
& Driverlog   & 88 & 74\% & 33\% & 48\% & 33\% \\
& Rovers      & 67 & 85\% & 43\% & 69\% & 49\% \\
& Satellite   & 100 & 73\% & 1\% & 52\% & 7\% \\
\midrule

\multirow{1}{*}{Mistral-Small-24B}
& Blocksworld & 99 & 50\% & 3\% & 18\% & 9\% \\
% & Satellite   & 39 & 46\% & 0\% & 24\% & 0\% \\
% \midrule
\bottomrule
\end{tabular}

\caption{Partial Experiment \#1 results for \% of total/hidden predicates identified and correct instances/prompt.}
\label{identify-preds-SMALL}
\end{table*}
% \end{table}
% \twocolumn}
% \twocolumn

%% file: AnonymousSubmission/LaTeX/paper-parts/added-preds-table.tex
\begin{table*}[h]
\centering
\renewcommand{\arraystretch}{1.15}
\setlength{\tabcolsep}{6pt}

\begin{tabular}{llcccccc}
\toprule
\multirow{2}{*}{\textbf{Model}} & \multirow{2}{*}{\textbf{Domain}}
& \multicolumn{3}{c}{\textbf{Prompt 1}}
& \multicolumn{3}{c}{\textbf{Prompt 2}} \\ 
\cmidrule(lr){3-5} \cmidrule(lr){6-8}
& 
& \textbf{\# Inst.} & \textbf{\% Added Preds} & \textbf{Avg. \#}
& \textbf{\# Inst.} & \textbf{\% Added Preds} & \textbf{Avg. \#} \\ 
\midrule

\multirow{5}{*}{Gemma-4-26B}
& Blocksworld & 95 & 19\% & 1 & 97 & 18\% & 1 \\
& Childsnack  & 85 & 19\% & 2 & 85 & 20\% & 3 \\
& Driverlog   & 82 & 7\%  & 4 & 87 & 3\%  & 4 \\
& Rovers      & 63 & 22\% & 2 & 62 & 39\% & 2 \\
& Satellite   & 92 & 17\% & 1 & 93 & 16\% & 1 \\
\midrule

\multirow{5}{*}{Qwen3.5-27B}
& Blocksworld & 100 & 25\% & 1 & 100 & 25\% & 1 \\
& Childsnack  & 87  & 12\% & 5 & 87  & 10\% & 7 \\
& Driverlog   & 89  & 7\%  & 2 & 88  & 3\%  & 9 \\
& Rovers      & 66  & 5\%  & 1 & 67  & 6\%  & 1 \\
& Satellite   & 98  & 5\%  & 2 & 100 & 8\%  & 1 \\
\midrule

\multirow{1}{*}{Mistral-Small-24B}
& Blocksworld & 100 & 82\% & 4 & 99 & 68\% & 3 \\
% & Satellite   & 39  & 85\% & 6 & 39 & 54\% & 4 \\
\bottomrule

\end{tabular}
\caption{Percent of Experiment \#1 answers with extraneous goal predicates added and the average number.}
% \caption{Percent of answers which contained extraneous goal predicates and the average number added for each model when it was asked to identify what a human's true goals were.}
% \caption{Percent of instances with extraneous predicates added to each model's answer about what a human's true goals were.}
% \caption{Percent of instances with extraneous predicates added to each model's final answer about what a human's true goals were.}
% \caption{Percent of instances with and the average number of extraneous predicates added to each model's final answers about what the human's true goals were.}
% what the robot believed the human's true goals were, and the average number added in cases where this happened.}
\label{num-preds-added-to-answer-exp1}
\end{table*}

%% file: AnonymousSubmission/LaTeX/paper-parts/SMALL_identify-preds-experiment2.tex
% \begin{table*}[h]

% \clearpage
% \afterpage{
% \onecolumn
\begin{table*}[t]
% \clearpage
% \begin{table}[t]
\centering
\renewcommand{\arraystretch}{1.15}
\setlength{\tabcolsep}{6pt}

% ================= TOP TABLE =================
\begin{tabular}{llccccc}
\toprule
\multirow{2}{*}{\textbf{Model}} & \multirow{2}{*}{\textbf{Domain}}
& \multicolumn{5}{c}{\textbf{Prompt 5.1}} \\ 
\cmidrule(lr){3-7}
&
& \textbf{\# Inst.} & \textbf{All} & \textbf{All \% Correct} & \textbf{MissH} & \textbf{MissH \% Correct} \\ 
\midrule

\multirow{5}{*}{Gemma-4-26B}
& Blocksworld & 93 & 83\% & 42\% & 66\% & 55\% \\
& Childsnack  & 79 & 75\% & 18\% & 55\% & 32\% \\
& Driverlog   & 74 & 71\% & 32\% & 52\% & 42\% \\
& Rovers      & 61 & 76\% & 21\% & 60\% & 39\% \\
& Satellite   & 78 & 59\% & 8\% & 37\% & 9\% \\
\midrule

\multirow{5}{*}{Qwen3.5-27B}
& Blocksworld & 100 & 74\% & 27\% & 47\% & 35\% \\
& Childsnack  & 88 & 72\% & 13\% & 48\% & 22\% \\
& Driverlog   & 87 & 79\% & 31\% & 57\% & 45\% \\
& Rovers      & 65 & 84\% & 37\% & 70\% & 52\% \\
& Satellite   & 98 & 72\% & 11\% & 51\% & 16\% \\
\midrule

\multirow{1}{*}{Mistral-Small-24B}
& Blocksworld & 95 & 70\% & 7\% & 46\% & 33\% \\
% & Satellite   & xx & xx\% & xx\% & xx\% & xx\% \\
% \bottomrule
\end{tabular}

% \vspace{0.6cm}

% ================= BOTTOM TABLE =================
\begin{tabular}{llccccc}
\toprule
\multirow{2}{*}{\textbf{Model}} & \multirow{2}{*}{\textbf{Domain}}
& \multicolumn{5}{c}{\textbf{Prompt 5.2}} \\ 
\cmidrule(lr){3-7}
&
& \textbf{\# Inst.} & \textbf{All} & \textbf{All \% Correct} & \textbf{MissH} & \textbf{MissH \% Correct} \\ 
\midrule

\multirow{5}{*}{Gemma-4-26B}
& Blocksworld & 90 & 84\% & 47\% & 69\% & 54\% \\
& Childsnack  & 82 & 69\% & 22\% & 50\% & 28\% \\
& Driverlog   & 79 & 70\% & 30\% & 53\% & 39\% \\
& Rovers      & 59 & 78\% & 25\% & 58\% & 37\% \\
& Satellite   & 77 & 60\% & 10\% & 39\% & 10\% \\
\midrule

\multirow{5}{*}{Qwen3.5-27B}
& Blocksworld & 100 & 74\% & 28\% & 47\% & 34\% \\
& Childsnack  & 88 & 71\% & 11\% & 47\% & 21\% \\
& Driverlog   & 87 & 78\% & 36\% & 56\% & 46\% \\
& Rovers      & 65 & 83\% & 32\% & 66\% & 48\% \\
& Satellite   & 97 & 73\% & 11\% & 53\% & 19\% \\
\midrule

\multirow{1}{*}{Mistral-Small-24B}
& Blocksworld & 96 & 70\% & 15\% & 44\% & 32\% \\
% & Satellite   & xx & xx\% & xx\% & xx\% & xx\% \\
\bottomrule
\end{tabular}

\caption{Partial Experiment \#2 results for \% of total/hidden predicates identified and correct instances/prompt.} 
\label{identify-preds-SMALL-exp2}
\end{table*}
% \end{table}

%% file: AnonymousSubmission/LaTeX/6-Discussion.tex
\section{Discussion} 
In general, Qwen was the strongest performing model, followed by Gemma and then Mistral. This could have been due to differences in reasoning ability, architecture density, context window sizes, or some combination thereof. Additionally, Qwen also asked slightly more questions about the robot's goals given the human's plan/inputted models than the other LLMs, which might have improved its performance. Overall, our findings show that Prompt 2 performed the worst, while Prompt 1 slightly outperformed Prompt 5.2, followed by Prompt 5.1. However, the success of these respective experiments seemed to be largely domain dependent, especially for Blocksworld. In particular, Prompt 5.2 achieved the best \% Correct results in 80\% of the domains tested with Gemma, but only in 20\% involving Qwen, and none with Mistral. This trend was also mostly reflected in the number of All and MissH predicates that were identified. 

One consistent trend across all models was that they had more difficulty identifying MissH predicates than getting these instances correct, and an easier time identifying the All predicates but more difficulty getting these instances correct. This highlights what may have been a potentially misleading statistic in the All category, as many of these answers only included the originally provided robot goals. This means a model may have achieved a high identification rate despite not solving anything. For example, a human could have had four goals but the model could have answered with the two already known to it, yielding a 50\% identification rate. In contrast, MissH predicates were never given to the models and therefore only ever reasoned about. Thus, while the All statistic is important, it should be interpreted with caution and the MissH category possibly treated as more revealing.

A central motivation for this work was analyzing whether LLMs could reason about a human's true goals given incomplete information. In Experiment \#2 Deepseek commonly answered questions about these by explicitly naming goal predicates. While this improved LLM performance in some cases, it may have partially negated the experiment's original purpose because, when these predicates were given, the model then did not have to reason about them. Conversely, Deepseek also sometimes answered with incorrect or misleading information. While beyond the scope of this work, quantifying the frequency of these issues and their downstream effects warrants further investigation. For example, examining what factors led to some answers being correct in Experiment \#1 but not in Experiment \#2. Or in which domains answers with incorrect information were most prevalent, given that these facts were meant to be beneficial.

%% file: AnonymousSubmission/LaTeX/paper-parts/2-related-work.tex
\section{Related Work}
The value alignment problem seeks to ensure that an agent's behavior and specified objectives align with the true intents of its user \cite{HadfieldValueAlignment,ChallengesValueAlignment}. Previously, most work has examined this in the context of decision-theoretic settings focused on the complexity of a human's objective specification mechanism \cite{leike2018,InverseRewardDesign}, or from a safety perspective where restrictions are placed on an agent or its environment to prevent behaviors that may produce unintended side effects \cite{AISafetyGridWorlds,GoalRecognitionDesign}. However, such work ignores a key cause of value misalignment: an agent's true behavior differing from what a human expected due to their incorrect beliefs about it.

Though research on this topic remains limited, some work on intent alignment \cite{IntentAlignment} has begun exploring the ways in which a user's instructions or objective specification can create misalignment. In \cite{Mechergui2024ExpectationAH}, this problem is modeled as a Markov Decision Process (MDP) and theory of mind is used to detect and handle a user's misspecified objectives. In \cite{ReducingGSD} environment design \cite{DesigningEnvironments} is used to modify an agent's surroundings such that it performs behaviors automatically aligned with a user's expectations. In \cite{Mechergui_Sreedharan_2024} the value alignment problem is reformulated to identify knowledge asymmetries arising from mismatched human-robot models. This information is then leveraged to determine a user's true objectives and satisfy them to the extent possible. Here, an optimal plan is found in the robot model which includes actions for making minimal queries to the user about specific goals likely missing from their specification and achieving all human goals. In our work, we adapted this formulation to work with LLMs and then evaluated how well these models were at identifying a user's true goals first in a zero-shot setting, and then after asking another LLM questions about them. For our setup, we were only concerned with identifying the human's goal predicates, not figuring out what agent plan would achieve them or if it was valid in the agent's model.

We motivated this work based on the observed fact that LLMs seldom ask questions about user requests, despite their frequent ambiguity and the value of such information. To induce this, past worked has explored prompting LLMs to ask clarifying questions \cite{elicitinghumanpreferences,lin-etal-2024-decision} and more recently enabled them to autonomously do this on their own \cite{ModelingFutureConvos}. To optimize query usefulness, others have used LLMs to determine what questions have the highest expected information gain \cite{looselipssinkships} or combined LLMs with other complementary methods to produce maximally informative questions \cite{BayesianPreference}. Conversely, LLMs have also been used to extract user goals from natural language conversations, avoiding questions altogether \cite{GoalInferenceOpenEndedDialog}.

%% file: AnonymousSubmission/LaTeX/8-conclusion-new.tex
\section{Future Work \& Conclusion}
In this paper, we focused on a common but often overlooked cause of value misalignment: a human misspecifying their goals due to incorrect beliefs about an agent's model. We then explored a derivative of this problem involving LLMs: whether asking clarifying questions about a human's goals improved a model's ability to accurately identify them. To evaluate this, we ran two experiments; One where each LLM was prompted to determine all user goals given misaligned human-robot models and a human plan, and another where they did the same thing but only after asking questions about them. We found that LLMs rarely ask questions unprompted and are poor at identifying goal predicates given incomplete and conflicting information. Further, when unsure about user intent, such models commonly add extra goal predicates to their answers. While our results do not conclusively show that asking questions always improves model performance, they do suggest that under certain conditions, their impact is significant. Especially for certain domains, and in terms of how well most models performed with respect to certain prompts.

In future work, we hope to expand upon the experiments presented here by performing a more comprehensive evaluation including a benchmark dataset, paid models, and other prompting techniques/styles. To avoid timeout issues and context window constraints, and to see how well certain models handle abstraction, we also hope to test on lifted and anonymized domains (i.e. domains with all object/predicate names replaced with alphanumeric strings) in addition to grounded domains. Additionally, aside from improving this study's pipeline and experiments, fine-tuning has been shown to improve model performance \cite{ACPBench}. Thus, we plan to see if this improves the average model identification accuracies presented in this paper.

%% file: AnonymousSubmission/LaTeX/paper-parts/7-appendix.tex
\section{Appendix: Do You Even Know What I Want? \\ Improving LLMs Ability to Detect Hidden User Intent} 

\vspace{2em} 

The supplementary file includes all information not included in Experiments \#1 and \#2 from the main paper. First, statistics for the grounded domains used in our experiments are presented. Next, the prompts used and the full predicate identification results/percent of correct instances achieved in Experiment \#1 are shown. Finally, the prompts, type/frequency of questions asked by each LLM, and the full predicate identification results/percent of correct instances achieved in Experiment \#2 are shown.

\vspace{2em} 

\section{Grounded Domain Statistics}

\input{AnonymousSubmission/LaTeX/paper-parts/domain-facts-table}

Table~\ref{domain-stats} shows statistics for each grounded domain tested in our experiments. The first two columns show the average number of predicates and actions in each grounded domain. The HGoals column corresponds to the total number of true human goals included in their problem file. For example, pay rent bill, use credit card, and schedule payment for end of month. The RGoals column represents the total number of goals in the robot's problem file, which is a subset of the human's true goals. For instance, pay rent bill. HPlan is the average number of actions the human believes the robot will execute to achieve their full set of goals (hidden + originally given). Or, more specifically, the average plan length across all optimal plans generated via the grounded human models with randomly added initial state predicates. RPlan is the average number of actions the robot will execute in true reality to achieve a subset of the human's true goals (i.e. only the goals originally given to it). Or, similar to before, the average plan length across all optimal plans generated via the grounded robot models, with a random number of goal predicates removed. As mentioned in the main paper, the number of predicates/actions in each grounded domain didn't seem to impact the percentage of correct instances achieved for the All and MissH \% Correct columns. However, the number of human versus robot goals did seem to have an effect, specifically for the Satellite domain. 

\vspace{3em}

\section{EXPERIMENT \#1}

\vspace{1em} 

\subsection{A. Prompts}
We begin this section by presenting Prompts 1 and 2 which were used to directly ask Gemma-4-26B, Llama-3.3-70B-Instruct, Mistral-Small-24B-Instruct, and Qwen3.5-27B to identify a human's true underlying goals given mismatched human-robot models and the human's plan, just using slightly different language. In response, we wanted to see if the LLMs would ask any questions unprompted about the misaligned human-robot models and human plan given to them. Then irrespective of this, we wanted to see how well they could identify the human's true goals. 

\newpage
% \subsection{Prompt 1}
\begin{figure}[h]
\centering
{\Large \textbf{Prompt 1}}\\
\rule{\linewidth}{0.0pt}
\makebox[\linewidth][c]{%
\begin{tcolorbox}[
width=\dimexpr\linewidth+2cm\relax,
colback=gray!10,
colframe=black,
boxrule=0.3pt,
arc=0pt
]
\scriptsize\ttfamily
\large

There’s a human who wants a robot to achieve some goals for them. Their models are as follows: \\ \\

\textbf{Human problem file:} \\  
\{\{human\_problem\}\} \\ \\

\textbf{Robot problem file:} \\ 
\{\{robot\_problem\}\} \\ \\

\textbf{Human/Robot domain file:} \\  
\{\{domain\_file\}\} \\ \\

Here’s the plan the human has given the robot to follow: 
\{\{valid\_plan\_output\}\} \\ \\

Achieving all these goals is the top priority but the goals the human has in their mind may be different than what has been specified. What do you think all of their goals could be? \\ \\

You should include all goal predicates from the robot problem file as well as whatever missing goals you think the human wanted. Wrap your final answer in tic marks, for example, like this ```I love the ocean```.

\end{tcolorbox}
}
% \caption{Prompt 1 from Experiment \#1.}
\caption{Prompt 1 from Experiment \#1 which asked a model to directly identify a human's true underlying goals given the provided information.}
\label{prompt-1}
\end{figure}

\newpage
% \subsection{Prompt 2}
\begin{figure}[h]
\centering
{\Large \textbf{Prompt 2}}\\
\rule{\linewidth}{0.0pt}
\makebox[\linewidth][c]{%
\begin{tcolorbox}[
width=\dimexpr\linewidth+2cm\relax,
colback=gray!10,
colframe=black,
boxrule=0.3pt,
arc=0pt
]
\scriptsize\ttfamily
\large

There’s a human who wants a robot to achieve some goals for them. Their models are as follows: \\ \\

\textbf{Human problem file:} \\  
\{\{human\_problem\}\} \\ \\

\textbf{Robot problem file:} \\ 
\{\{robot\_problem\}\} \\ \\

\textbf{Human/Robot domain file:} \\  
\{\{domain\_file\}\} \\ \\

To achieve them, the plan the human thinks the robot should follow is:  
\{\{valid\_plan\_output\}\} \\ \\

The robot’s goals are underspecified and are a subset of the human’s goals. Some goals might be unachievable. What do you think all of the human's true goals could be? \\ \\

You should include all goal predicates from the robot problem file as well as whatever missing goals you think the human wanted. Wrap your final answer in tic marks, for example, like this ```I love the ocean```.

\end{tcolorbox}
}
\caption{Prompt 2 from Experiment \#1 which asked a model to directly identify a human's true underlying goals given the provided information.}
\label{prompt-2}

\end{figure}

\vspace{2em} 

\subsection{B. Predicate Identification Accuracies}

In this experiment, we found the LLMs did not tend to ask questions unprompted, and that Prompt 1 far outperformed Prompt 2. We present these findings on the next page, specifically showing the achieved identification accuracy for the models, domains, and prompts used in Experiment \#1. In the presented tables, \# Inst. refers to the number of completed instances on which all subsequent metrics for each prompt are based. The All column captures what percent of all possible predicates (those given and withheld) that a model across all problem instances was able to identify. The MissH column captures what percent of predicates hidden from a model across all instances it was able to identify. All $\%$ Correct and MissH $\%$ Correct refer to the percent of completely correct instances for each scoring criteria that this translated into.

\newpage
\input{AnonymousSubmission/LaTeX/paper-parts/identify-preds-table}
\FloatBarrier

\section{EXPERIMENT \#2}

\vspace{1em} 

\subsection{A. Prompts}
We begin this section by presenting Prompts 3, 4, 5.1 and 5.2 which were used to test each LLMs ability to identify, ask, or answer questions about a human's true goals. Recall that Prompt 3 was used to have Gemma-4-26B, Mistral-Small-24B-Instruct, and Qwen3.5-27B ask questions about a human's true underlying goals given mismatched human-robot models and the human's plan. Prompt 4 then fed these questions to Deepseek-R1-Distill-Llama-70B along with the human's true goals and asked it to answer them. These responses and all previously inputted information was then fed into Prompts 5.1 and 5.2. Using slightly different language, these directly asked Gemma-4-26B, Mistral-Small-24B-Instruct, and Qwen3.5-27B to identify a human's true underlying goals. Here, we wanted to see if directly prompting each LLM to ask questions about a human's true goals would improve the average model identification accuracy, and the type/frequency/usefulness of the questions being asked.

\vspace{3em} 

% \newpage
% \subsection{Prompt 3}
\begin{figure}[h]
\centering
{\Large \textbf{Prompt 3}}\\
\rule{\linewidth}{0.0pt}
\makebox[\linewidth][c]{%
\begin{tcolorbox}[
width=\dimexpr\linewidth+2cm\relax,
colback=gray!10,
colframe=black,
boxrule=0.3pt,
arc=0pt
]
\scriptsize\ttfamily
\large

There’s a human who wants a robot to achieve some goals for them. Their models are as follows: \\ \\

\textbf{Human problem file:} \\  
\{\{human\_problem\}\} \\ \\

\textbf{Robot problem file:} \\ 
\{\{robot\_problem\}\} \\ \\

\textbf{Human/Robot domain file:} \\  
\{\{domain\_file\}\} \\ \\

To achieve them, the plan the human thinks the robot should follow is:  
\{\{valid\_plan\_output\}\} \\ \\

The robot’s goals are underspecified and are a subset of the human’s goals. Some goals might be unachievable. What questions should we ask to figure out what the human's true goals are? \\ \\

Wrap your final answer in tic marks, for example, like this ```I love the ocean```.

\end{tcolorbox}
}
\caption{Prompt 3 from Experiment \#2 which asked a model to ask questions about a human's true underlying goals given the provided information.}
\label{prompt-3}
\end{figure}

\newpage
% \subsection{Prompt 4}
\begin{figure}[h]
\centering
{\Large \textbf{Prompt 4}}\\
\rule{\linewidth}{0.0pt}
\makebox[\linewidth][c]{%
\begin{tcolorbox}[
width=\dimexpr\linewidth+2cm\relax,
colback=gray!10,
colframe=black,
boxrule=0.3pt,
arc=0pt
]
\scriptsize\ttfamily
\large

There’s a human who wants a robot to achieve some goals for them. But the robot's goals are underspecified and are a subset of the human’s goals. \\ \\

The human's goals are as follows: \\ \\
 
\{\{human\_goals\}\} \\ \\

To better understand what the human wants the robot has the following questions: \\ \\ 
 
\{\{robot\_questions\}\} \\ \\

Answer all of the robot's questions to help it satisfy what the human wants. \\ \\ 

Wrap your final answer in tic marks, for example, like this ```I love the ocean```. \\ \\

You are required to do this and only the answers you include inside the tic marks will be viewed.

\end{tcolorbox}
}
\caption{Prompt 4 from Experiment \#2 which asked Deepseek-R1-Distill-Llama-70B to answer an LLM's questions from Prompt 3 given the human's true goals. Here, Deepseek was meant to approximate a human with full knowledge of the task answering questions about it.}
\label{prompt-4}

\end{figure}

\newpage
% \subsection{Prompt 5.1}
\begin{figure}[h]
\centering
{\Large \textbf{Prompt 5.1}}\\
\rule{\linewidth}{0.0pt}
\makebox[\linewidth][c]{%
\begin{tcolorbox}[
width=\dimexpr\linewidth+2cm\relax,
colback=gray!10,
colframe=black,
boxrule=0.3pt,
arc=0pt
]
\scriptsize\ttfamily
\large

There’s a human who wants a robot to achieve some goals for them. Their models are as follows: \\ \\

\textbf{Human problem file:} \\  
\{\{human\_problem\}\} \\ \\

\textbf{Robot problem file:} \\ 
\{\{robot\_problem\}\} \\ \\

\textbf{Human/Robot domain file:} \\  
\{\{domain\_file\}\} \\ \\

Here’s the plan the human has given the robot to follow: \\
\{\{valid\_plan\_output\}\} \\ \\

Answers to questions you had about all of this are: \\
\{\{deepseek\_answers\}\} \\ \\  

Achieving all these goals is the top priority but the goals the human has in their mind may be different than what has been specified. What do you think all of their goals could be? \\ \\

You should include all goal predicates from the robot problem file as well as whatever missing goals you think the human wanted. Wrap your final answer in tic marks, for example, like this ```I love the ocean```.

\end{tcolorbox}
}
\caption{Prompt 5.1 from Experiment \#2 which asked a model to directly identify a human's true underlying goals given the provided model/plan information and answers from Deepseek about them.}
\label{prompt-5.1}
\end{figure}

\newpage
% \subsection{Prompt 5.2}
\begin{figure}[h]
\centering
{\Large \textbf{Prompt 5.2}}\\
\rule{\linewidth}{0.0pt}
\makebox[\linewidth][c]{%
\begin{tcolorbox}[
width=\dimexpr\linewidth+2cm\relax,
colback=gray!10,
colframe=black,
boxrule=0.3pt,
arc=0pt
]
\scriptsize\ttfamily
\large

There’s a human who wants a robot to achieve some goals for them. Their models are as follows: \\ \\

\textbf{Human problem file:} \\  
\{\{human\_problem\}\} \\ \\

\textbf{Robot problem file:} \\ 
\{\{robot\_problem\}\} \\ \\

\textbf{Human/Robot domain file:} \\  
\{\{domain\_file\}\} \\ \\

To achieve them, the plan the human thinks the robot should follow is: \\ 
\{\{valid\_plan\_output\}\} \\ \\

Answers to questions you had about all of this are: \\
\{\{deepseek\_answers\}\} \\ \\ 

The robot’s goals are underspecified and are a subset of the human’s goals. Some goals might be unachievable. What do you think all of the human's true goals could be? \\ \\

You should include all goal predicates from the robot problem file as well as whatever missing goals you think the human wanted. Wrap your final answer in tic marks, for example, like this ```I love the ocean```.

\end{tcolorbox}
}
\caption{Prompt 5.2 from Experiment \#2 which asked a model to directly identify a human's true underlying goals given the provided model/plan information and answers from Deepseek about them.}
\label{prompt-5.2}
\end{figure}

\vspace{2em} 

\subsection{B. Type \& Frequency of Questions Asked in Response to Prompt 3}

The next section presents additional details about the questions asked by the Gemma-4-26B, Mistral-Small-24B-Instruct, and Qwen3.5-27B models in response to Prompt 3. Specifically, Table~\ref{questions-asked} (see below) shows the average number of questions asked by each model in response to Prompt 3 (shown above), where \# Inst. refers to the number of completed instances on which these metrics are based, and Avg. Q shows the average number of questions asked by each model across domains. Table~\ref{fig:question-frequency} (see below) then shows the frequency and type of questions asked. These fell into two main categories, goal clarification and plan versus goal alignment. The key takeaways from this data was (1) that Mistral averaged the most questions with 6, followed by Qwen with 4, and Gemma with 2, (2) most questions were about understanding how the misaligned human-robot models/plans aligned with the humans goals instead of purely just clarifying goals, and (3) this was most obvious for the Qwen model which focused less than 20\% of its questions on purely clarifying goals.

\newpage

\section{Average Number of Questions Asked in Response to Prompt 3} 
\input{AnonymousSubmission/LaTeX/paper-parts/questions-asked-table}

\vspace{0.3em} 
% Table~\ref{questions-asked} shows the average number of questions asked by each model in response to Prompt 3 (shown above). \# Inst. refers to the number of completed instances on which these metrics are based. Avg. Q shows the average number of questions asked by each model across domains. Here, Mistral averaged the most questions with 6, followed by Qwen with 4, and Gemma with 2. 

\vspace{1.5em} 
\section{Question Type Frequency in Response to Prompt 3}
\begin{figure*}[h]
% \begin{figure*}[t]
    \centering
    % Top row
    \begin{minipage}[t]{0.49\textwidth}
        \centering
        \includegraphics[width=\linewidth]{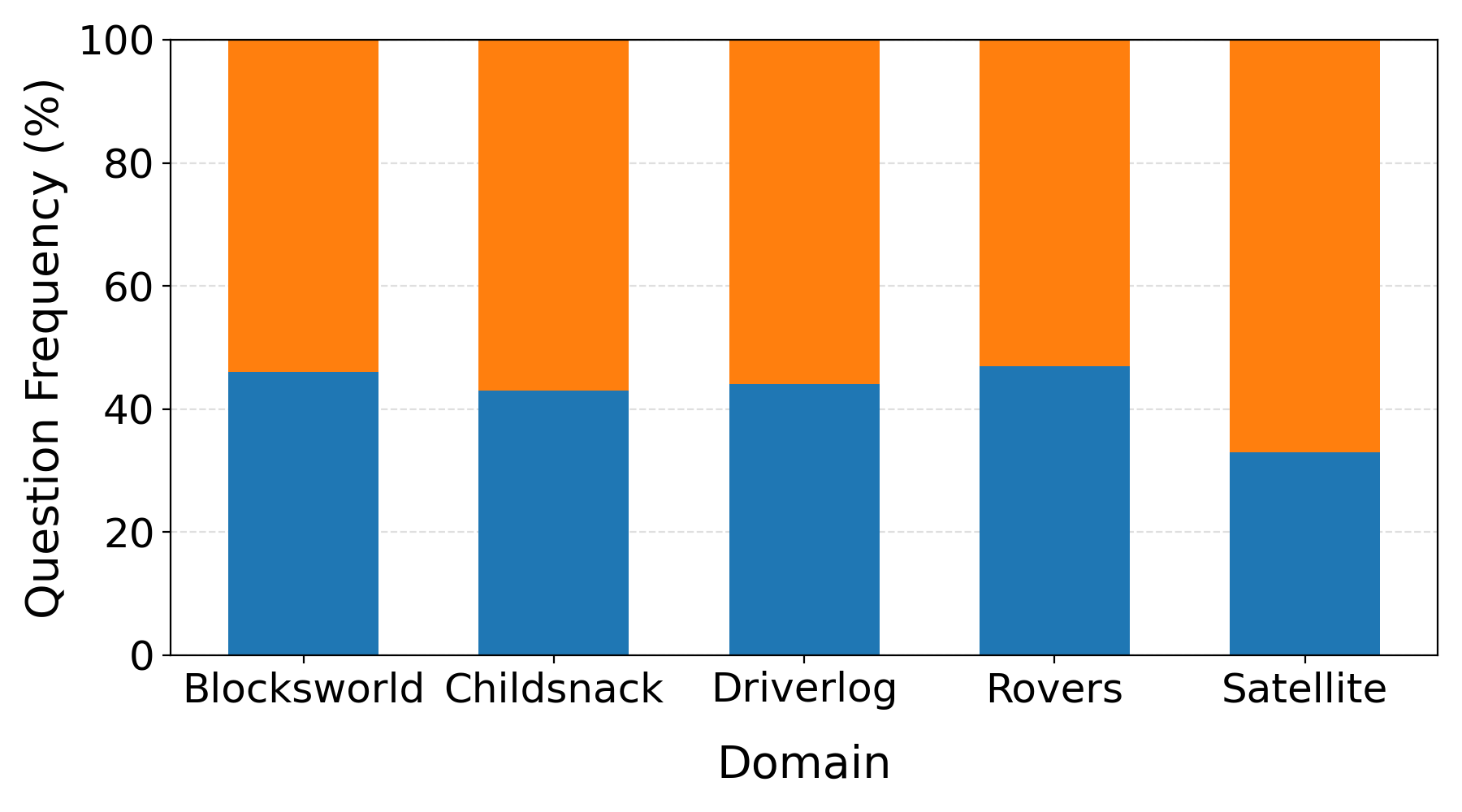}
        \caption*{(a) Gemma-4-26B}
    \end{minipage}
    \hfill
    \begin{minipage}[t]{0.49\textwidth}
        \centering
        \includegraphics[width=\linewidth]{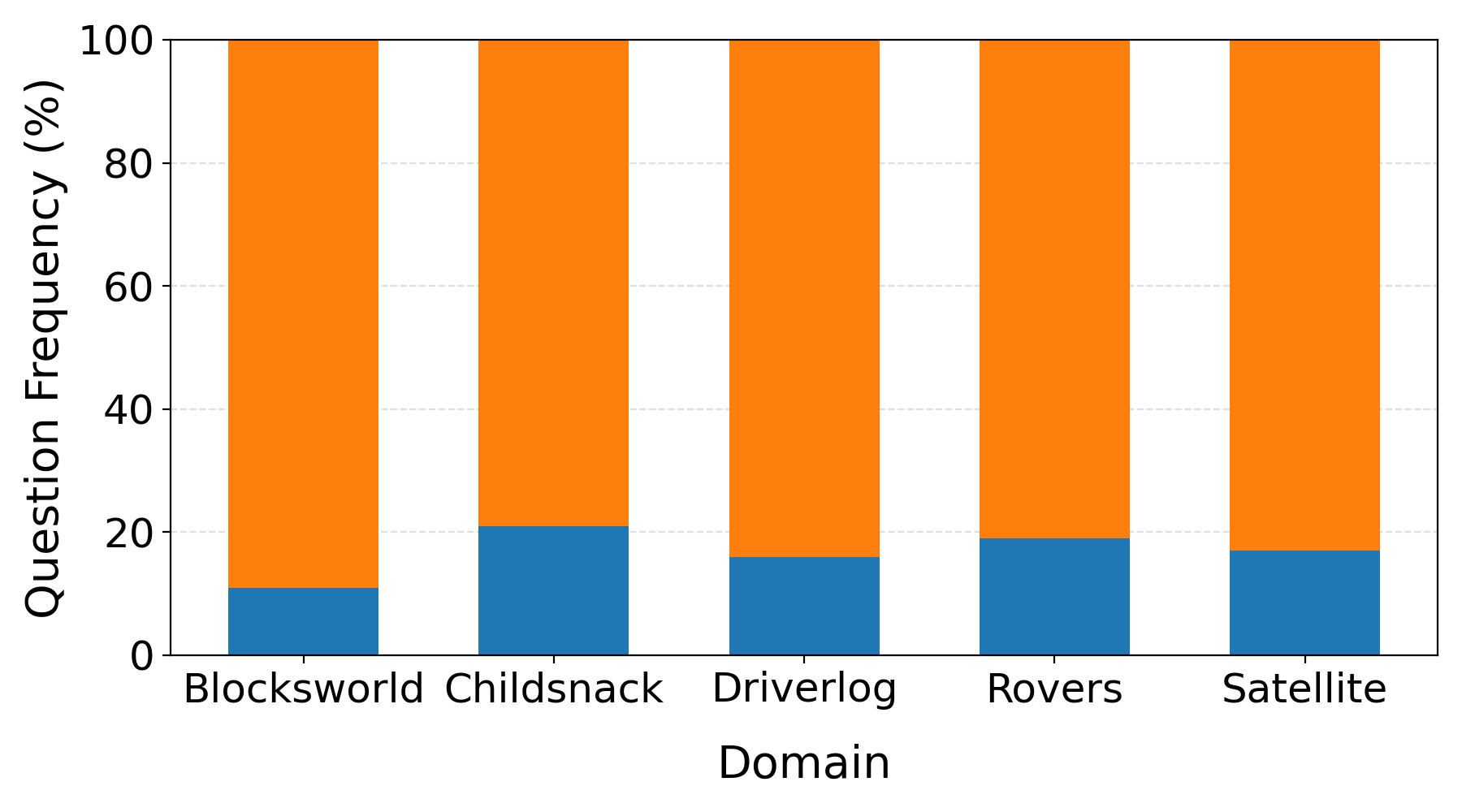}
        \caption*{(b) Qwen3.5-27B}
    \end{minipage}
    \vspace{6pt}
    % Bottom row: Mistral + legend
    \begin{minipage}[c]{0.49\textwidth}
        \centering
        \includegraphics[width=\linewidth]{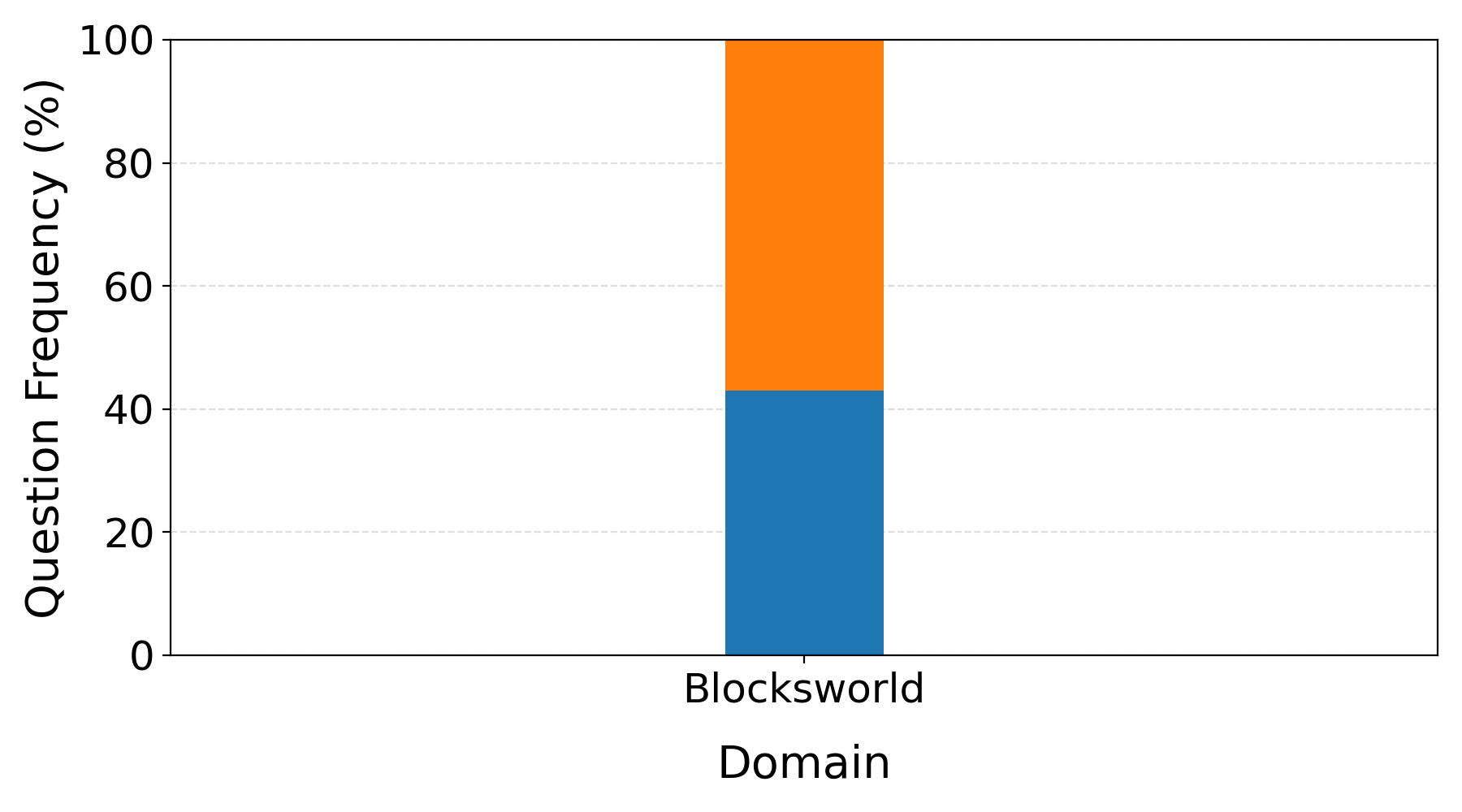}
        \caption*{(c) Mistral-Small-24B}
    \end{minipage}
    \hfill
    \begin{minipage}[c]{0.49\textwidth}
    \centering
    \raisebox{1.8cm}{
        \includegraphics[width=0.90\linewidth]{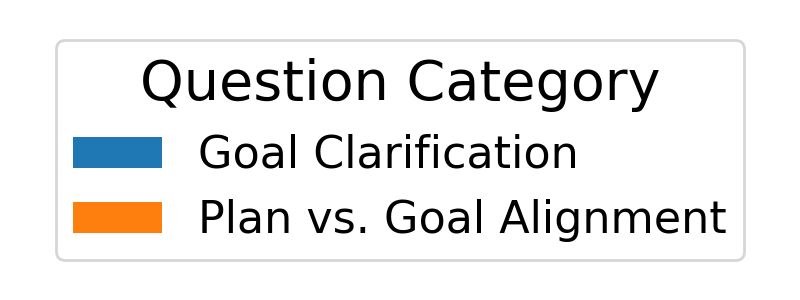}
    }
    \end{minipage}
    \caption{Question type frequency in response to Prompt 3 for each model by domain.}
    \label{fig:question-frequency}
\end{figure*}

% \section{Experiment \#2: Predicate Identification Results}

\subsection{C. Predicate Identification Accuracies}
In Experiment \#2, we found that asking questions for some domains (specifically Blocksworld) greatly improved the average model identification accuracy. Here, we present these findings, specifically showing the achieved identification accuracy for the models, domains, and prompts used in this experiment. Like before, in the presented tables, \# Inst. refers to the number of completed instances on which all subsequent metrics for each prompt are based. The All column captures what percent of all possible predicates (those given and withheld) that a model across all problem instances was able to identify. The MissH column captures what percent of predicates hidden from a model across all instances it was able to identify. All $\%$ Correct and MissH $\%$ Correct refer to the percent of completely correct instances for each scoring criteria that this translated into.

\vspace{3em} 

\input{AnonymousSubmission/LaTeX/paper-parts/identify-preds-experiment2}

%% file: AnonymousSubmission/LaTeX/paper-parts/domain-facts-table.tex
\begin{table*}[h]
\centering
%\tiny
\begin{tabular}{|c|c|c|c|c|c|c|c|c|c|c|}
\hline
\textbf{Domain} & \textbf{\# Predicates} & \textbf{\# Actions} & \textbf{HGoals} & \textbf{HPlan} & \textbf{RGoals} & \textbf{RPlan}\\ \hline
Blocksworld & 60 & 93 & 4 & 8 & 2 & 6 \\ \hline
Childsnack & 102 & 1122 & 4 & 8 & 2 & 5 \\ \hline
% Depots? & xx & xx & xx & xx & xx & xx \\ \hline
Driverlog & 471 & 1216 & 4 & 11 & 2 & 8\\ \hline
% Floortile? & xx & xx & xx & xx & xx & xx\\ \hline
Rovers & 194 & 1071 & 4 & 9 & 2 & 6\\ \hline
Satellite & 116 & 536 & 5 & 9 & 2 & 6\\ \hline
\end{tabular}
\caption{Statistics for each grounded domain/problem file, and human/robot plan.}
\label{domain-stats}
\end{table*}

%% file: AnonymousSubmission/LaTeX/paper-parts/identify-preds-table.tex
% \clearpage
% \afterpage{
% \onecolumn
% \begin{table*}[t]

\clearpage
\begin{table}[t]
% \begin{table}[p]
\centering
\renewcommand{\arraystretch}{1.15}
\setlength{\tabcolsep}{6pt}

% -------- Prompt 1 --------
\begin{tabular}{llccccc}
\toprule
\multirow{2}{*}{\textbf{Model}} & \multirow{2}{*}{\textbf{Domain}}
& \multicolumn{5}{c}{\textbf{Prompt 1}} \\ 
\cmidrule(lr){3-7}
& 
& \textbf{\# Inst.} & \textbf{All} & \textbf{All \% Correct} & \textbf{MissH} & \textbf{MissH \% Correct} \\ 
\midrule

\multirow{5}{*}{Gemma-4-26B}
& Blocksworld & 95 & 65\% & 13\% & 33\% & 18\% \\
& Childsnack  & 85 & 71\% & 20\% & 49\% & 24\% \\
& Driverlog   & 82 & 63\% & 31\% & 37\% & 32\% \\
& Rovers      & 63 & 62\% & 22\% & 48\% & 29\% \\
& Satellite   & 92 & 47\% & 4\% & 30\% & 8\% \\
\midrule

\multirow{5}{*}{Qwen3.5-27B}
& Blocksworld & 100 & 72\% & 19\% & 44\% & 27\% \\
& Childsnack  & 87 & 74\% & 24\% & 52\% & 25\% \\
& Driverlog   & 89 & 82\% & 39\% & 66\% & 45\% \\
& Rovers      & 66 & 88\% & 52\% & 75\% & 56\% \\
& Satellite   & 98 & 74\% & 12\% & 55\% & 15\% \\
\midrule

\multirow{5}{*}{Mistral-Small-24B}
& Blocksworld & 100 & 63\% & 3\% & 33\% & 18\% \\
& Childsnack  & -- & -- & -- & -- & -- \\
& Driverlog   & 11 & 97\% & 18\% & 93\% & 91\% \\
& Rovers      & 6 & 87\% & 0\% & 73\% & 50\% \\
& Satellite   & 39 & 79\% & 3\% & 77\% & 49\% \\
\midrule

\multirow{5}{*}{Llama-3.1}
& Blocksworld & 14 & 71\% & 14\% & 59\% & 50\% \\
& Childsnack  & 1 & 33\% & 0\% & 0\% & 0\% \\
& Driverlog   & 1 & 0\% & 0\% & 0\% & 0\% \\
& Rovers      & 5 & 74\% & 0\% & 71\% & 60\% \\
& Satellite   & 13 & 58\% & 0\% & 42\% & 23\% \\
\bottomrule
\end{tabular}

\vspace{1em}

% -------- Prompt 2 --------
\begin{tabular}{llccccc}
\toprule
\multirow{2}{*}{\textbf{Model}} & \multirow{2}{*}{\textbf{Domain}}
& \multicolumn{5}{c}{\textbf{Prompt 2}} \\ 
\cmidrule(lr){3-7}
& 
& \textbf{\# Inst.} & \textbf{All} & \textbf{All \% Correct} & \textbf{MissH} & \textbf{MissH \% Correct} \\ 
\midrule

\multirow{5}{*}{Gemma-4-26B}
& Blocksworld & 97 & 64\% & 11\% & 28\% & 16\% \\
& Childsnack  & 85 & 69\% & 17\% & 46\% & 20\% \\
& Driverlog   & 87 & 61\% & 16\% & 28\% & 16\% \\
& Rovers      & 62 & 70\% & 13\% & 46\% & 29\% \\
& Satellite   & 93 & 52\% & 1\% & 25\% & 3\% \\
\midrule

\multirow{5}{*}{Qwen3.5-27B}
& Blocksworld & 100 & 69\% & 18\% & 37\% & 22\% \\
& Childsnack  & 87 & 71\% & 20\% & 46\% & 21\% \\
& Driverlog   & 88 & 74\% & 33\% & 48\% & 33\% \\
& Rovers      & 67 & 85\% & 43\% & 69\% & 49\% \\
& Satellite   & 100 & 73\% & 1\% & 52\% & 7\% \\
\midrule

\multirow{5}{*}{Mistral-Small-24B}
& Blocksworld & 99 & 50\% & 3\% & 18\% & 9\% \\
& Childsnack  & -- & -- & -- & -- & -- \\
& Driverlog   & 11 & 78\% & 55\% & 43\% & 55\% \\
& Rovers      & 6 & 83\% & 17\% & 64\% & 33\% \\
& Satellite   & 39 & 46\% & 0\% & 24\% & 0\% \\
\midrule

\multirow{5}{*}{Llama-3.1}
& Blocksworld & 3 & 55\% & 0\% & 0\% & 0\% \\
& Childsnack  & 1 & 0\% & 0\% & 0\% & 0\% \\
& Driverlog   & 1 & 83\% & 0\% & 67\% & 0\% \\
& Rovers      & 1 & 75\% & 0\% & 67\% & 0\% \\
& Satellite   & 2 & 82\% & 0\% & 71\% & 0\% \\
\bottomrule
\end{tabular}

\caption{Full Experiment \#1 results showing the \% of total/hidden predicates identified and correct instances/prompt.}
% \caption{Experiment \#1 results showing accuracy across models, domains, and prompts specific to identifying predicates associated with a human's true goals. \# Inst. refers to the number of completed instances on which all subsequent metrics for each prompt are based. The All column captures what percent of all possible predicates (those given and withheld) that a model across all problem instances was able to identify. The MissH column captures what percent of predicates hidden from a model across all instances it was able to identify. All $\%$ Correct and MissH $\%$ Correct refer to the percent of completely correct instances which involved either all predicates or only those that were withheld but a part of the human's true goals.}
\label{identify-preds}
% \end{table*}
\end{table}

%% file: AnonymousSubmission/LaTeX/paper-parts/questions-asked-table.tex
% \begin{table}[!t]
% \centering
\vspace{1em} 
\begin{center}
\footnotesize
\setlength{\tabcolsep}{4pt}
\begin{tabular}{llcc}
\toprule
& & \multicolumn{2}{c}{\textbf{Prompt 3}} \\
\cmidrule(lr){3-4}
\textbf{Model} & \textbf{Domain} & \textbf{Inst.} & \textbf{Avg. Q} \\
\midrule
\multirow{5}{*}{Gemma-4-26B}
& Blocksworld & 98 & 2 \\ 
& Childsnack  & 80 & 3 \\ 
& Driverlog   & 82 & 2 \\ 
& Rovers      & 65 & 2 \\ 
& Satellite   & 87 & 2 \\ 
\midrule
\multirow{5}{*}{Qwen3.5-27B}
& Blocksworld & 100 & 4 \\ 
& Childsnack  & 88 & 4 \\ 
& Driverlog   & 89 & 4 \\ 
& Rovers      & 66 & 4 \\ 
& Satellite   & 100 & 4 \\ 
\midrule
\multirow{1}{*}{Mistral-Small-24B}
& Blocksworld & 61 & 6 \\ 
\bottomrule
\end{tabular}
% \caption{{Prompt 3 average \# of questions asked/model.}}
\captionof{table}{Prompt 3 average \# of questions asked/model.}
\label{questions-asked}
\end{center}
% \end{table}

%% file: AnonymousSubmission/LaTeX/paper-parts/identify-preds-experiment2.tex
% \begin{table*}[h]

% \clearpage
% \afterpage{
% \onecolumn
% \begin{table*}[t]
% \clearpage
\begin{table}[h]
\centering
\renewcommand{\arraystretch}{1.15}
\setlength{\tabcolsep}{6pt}

% ================= TOP TABLE =================
\begin{tabular}{llccccc}
\toprule
\multirow{2}{*}{\textbf{Model}} & \multirow{2}{*}{\textbf{Domain}}
& \multicolumn{5}{c}{\textbf{Prompt 5.1}} \\ 
\cmidrule(lr){3-7}
&
& \textbf{\# Inst.} & \textbf{All} & \textbf{All \% Correct} & \textbf{MissH} & \textbf{MissH \% Correct} \\ 
\midrule

\multirow{5}{*}{Gemma-4-26B}
& Blocksworld & 93 & 83\% & 42\% & 66\% & 55\% \\
& Childsnack  & 79 & 75\% & 18\% & 55\% & 32\% \\
& Driverlog   & 74 & 71\% & 32\% & 52\% & 42\% \\
& Rovers      & 61 & 76\% & 21\% & 60\% & 39\% \\
& Satellite   & 78 & 59\% & 8\% & 37\% & 9\% \\
\midrule

\multirow{5}{*}{Qwen3.5-27B}
& Blocksworld & 100 & 74\% & 27\% & 47\% & 35\% \\
& Childsnack  & 88 & 72\% & 13\% & 48\% & 22\% \\
& Driverlog   & 87 & 79\% & 31\% & 57\% & 45\% \\
& Rovers      & 65 & 84\% & 37\% & 70\% & 52\% \\
& Satellite   & 98 & 72\% & 11\% & 51\% & 16\% \\
\midrule

\multirow{1}{*}{Mistral-Small-24B}
& Blocksworld & 95 & 70\% & 7\% & 46\% & 33\% \\
% & Satellite   & xx & xx\% & xx\% & xx\% & xx\% \\
\bottomrule
\end{tabular}

\vspace{0.6cm}

% ================= BOTTOM TABLE =================
\begin{tabular}{llccccc}
\toprule
\multirow{2}{*}{\textbf{Model}} & \multirow{2}{*}{\textbf{Domain}}
& \multicolumn{5}{c}{\textbf{Prompt 5.2}} \\ 
\cmidrule(lr){3-7}
&
& \textbf{\# Inst.} & \textbf{All} & \textbf{All \% Correct} & \textbf{MissH} & \textbf{MissH \% Correct} \\ 
\midrule

\multirow{5}{*}{Gemma-4-26B}
& Blocksworld & 90 & 84\% & 47\% & 69\% & 54\% \\
& Childsnack  & 82 & 69\% & 22\% & 50\% & 28\% \\
& Driverlog   & 79 & 70\% & 30\% & 53\% & 39\% \\
& Rovers      & 59 & 78\% & 25\% & 58\% & 37\% \\
& Satellite   & 77 & 60\% & 10\% & 39\% & 10\% \\
\midrule

\multirow{5}{*}{Qwen3.5-27B}
& Blocksworld & 100 & 74\% & 28\% & 47\% & 34\% \\
& Childsnack  & 88 & 71\% & 11\% & 47\% & 21\% \\
& Driverlog   & 87 & 78\% & 36\% & 56\% & 46\% \\
& Rovers      & 65 & 83\% & 32\% & 66\% & 48\% \\
& Satellite   & 97 & 73\% & 11\% & 53\% & 19\% \\
\midrule

\multirow{1}{*}{Mistral-Small-24B}
& Blocksworld & 96 & 70\% & 15\% & 44\% & 32\% \\
% & Satellite   & xx & xx\% & xx\% & xx\% & xx\% \\
\bottomrule
\end{tabular}

\caption{Full Experiment \#2 results showing the \% of total/hidden predicates identified and correct instances/prompt.} 

% identified and All \% Correct what percent of correct problem instances this translated into. MissH is the percent of hidden human goals each model identified and MissH \% Correct what percent of correct instances this resulted in.}

% \caption{Experiment \#2 results for Prompt \#5.1 and \#5.2 where multi-state prompting was used. \#Inst. is the total number of completed instances. All is the total percent of all possible predicates that were identified and All \% Correct what percent of correct problem instances this translated into. MissH is the percent of hidden human goals each model identified and MissH \% Correct what percent of correct instances this resulted in.}

% The percent of hidden and total human goals identified by each model in Experiment \#2 for Prompt \#5.1 and \#5.2. }
\label{identify-preds-exp2}
% \end{table*}
\end{table}